# Deep Invertible Networks for EEG-based brain-signal decoding

*Robin Tibor Schirrmeister[1,2], Tonio Ball[1]*
July 14th, 2019

*In this manuscript, we investigate deep invertible networks for EEG-based brain signal decoding and find them to generate realistic EEG signals as well as classify novel signals above chance. Further ideas for their regularization towards better decoding accuracies are discussed.*

## Introduction

Deep-learning-based brain-signal decoding has recently achieved competitive accuracies compared with traditional feature-based decoding approaches. For example, they were used to decode movement-related EEG signals with accuracies at least as good as well-established movement-decoding approaches (Schirrmeister et al., 2017a) and been applied to error or event-related-based decoding (Lawhern et al., 2018, Völker et al., 2018) as well as automatic diagnosis of pathologies (Schirrmeister et. al, 2017b).

Still, there are several desirable things that are challenging to achieve with conventional deep-learning-based brain-signal decoding. Among them are:
- interpret the trained models
- integrate unlabelled data
- detect if the distribution of the EEG signals has changed since the network was trained

A principled approach that tackles these challenges are generative networks. Generative network in a class-conditional setting are here defined as networks that learn the joint likelihood of any input and label combination. As such, these methods can integrate unlabelled data for semi-supervised learning by optimizing the conditional likelihood for the datapoints with labels and the unconditional likelihood for the datapoints without labels (Izmailov et al., 2019)

A recent promising generative network approach are invertible networks (Dinh et al., 2014, Dinh et al., 2016) They can learn probability distributions in the input space by keeping track of how much the input-output mapping defined by the invertible network squeezes or expands volume and using a predefined output distribution such as a gaussian distribution. One property that makes them particularly interesting for high-dimensional data that often arises in the medical domain is their low memory usage: Due to their invertibility, they have a constant memory usage with regards to the depth of the network in contrast to a linearly increasing memory usage for conventional deep networks. This allows to train deeper networks even on very high-dimensional data.

[1] Translational Neurotechnology Lab, Medical Center - University of Freiburg, Germany
[2] Machine Learning Lab, University of Freiburg, Germany. Correspondence to: Robin Tibor Schirrmeister<robin.schirrmeister@uniklinik-freiburg.de>.

In this preliminary work, we apply invertible networks to the task of EEG-based brain-signal decoding. We develop an invertible architecture that can create visually indistinguishable synthetic EEG signals and simultaneously classify novel EEG signals. Apart from visual analysis, we also validated the quality of the generated signals by showing that a conventional deep neural network classifies them largely into the correct classes and that a conventional deep neural network trained purely on these generated signals classifies real EEG signals better than chance. We present several ideas how to regularize the invertible classification networks to further close the still-existing gap between their decoding accuracies and those of conventional convolutional networks.

# Methods

## Invertible Networks

Invertible networks (Dinh et al., 2014, Dinh et al., 2016) are networks where each building block of the network is invertible by design. One popular such building block are additive or affine coupling layers that work by splitting the input x into disjoint parts x1,x2. For the additive coupling block, it proceeds to compute its output as:

$y_1 = F(x_2) + x_1$
$y_2 = G(y_1) + x_2$

This can be easily inverted by:

$x_2 = y_2 - G(y_1)$
$x_1 = y_1 - F(x_2)$

One key property of these additive coupling blocks is that they do not change any volume from input to output. Therefore, given a prior probability distribution in the output space (e.g., a gaussian distribution) and a chain of invertible blocks, if you compute the likelihood values at each point in the input space by using the prior in the output space and the inverse of the chain of additive blocks, the likelihood values still integrate to 1 in the input space, and therefore still constitute a proper probability distribution. For other types of transformations, to ensure you still get a proper probability distribution in the input space, you have to keep track of the volume changes at any point from input to output and account for it by the changes of variable formula.

The probability $p_X(x)$ of point $x$ in the input space $X$ can then be computed from the mapping function $f$ (e.g., as implemented by a neural network) and the prior $p_H$ in output space as such:
Us:
$p_X(x) = p_H(f(x)) \cdot |det\frac{\partial f(x)}{\partial x}|$, where $\frac{\partial f(x)}{\partial x}$ is the jacobian of the mapping function $f$ with regards to input $x$. For an arbitrary network, this jacobian is very expensive to compute.

Since the additive coupling blocks do not change volume, they are very easy to train by maximum likelihood of the data under the invertible network and the given prior by directly maximizing $p_H(f(x))$. As the prior distribution, we take a class-conditional uncorrelated gaussian distribution, and also optimize the means and standard deviations of each class-conditional gaussian distribution jointly with the network parameters. This is mathematically identical to the scaling layer introduced in Dinh et al. (2014) and an additional bias layer, with both of them having independent parameters for each class. We find this to optimize fairly robustly, in contrast to a recent work that attempted to optimize the entire covariance matrix of the class-conditional gaussian distributions.

There are also other invertible building blocks that allow for an easy computation of these volume change terms. Other building blocks can be more expressive than pure additive coupling blocks. Additive coupling blocks are inherently constrained in what they can represent - for example, additive coupling blocks will always transform a uniform distribution into another uniform distribution, just with potentially different support. Nevertheless, additive coupling blocks still often achieve decent performance and are potentially more stable to train, therefore we used them in this first attempt to create an invertible network on EEG signals.

One issue with maximum likelihood training for invertible networks is that you normally only have a finite training set which has a volume of 0. Therefore the network may overfit by assigning arbitrarily high likelihoods to the training data points. One way to alleviate this is by adding a small uniform dequantization noise to the data, originally motivated to undo the quantization of images to 256 discrete color values (Theis et al., 2015). Beyond that, noise may also be added in different ways to optimize or regularize the invertible network further (Ardizzone et al., 2019; Ho et al., 2019).

## Optimal transport

As an alternative to maximum likelihood training, we also explore optimal transport based optimization. The optimal transport distance measures the cost of morphing one distribution into another one as follows:
1. Define a distance function on the support of both distributions, for example the euclidean distance.

2. Find a weight for any pair of points $x, y$ where x is from the support of distribution $P$ and $y$ is from the support of distribution $Q$ such that: (1) the sum of the weights at each point $x$ equals the probability $p(x)$ and at each point $y$ equals the probability $q(y)$. (2) The sum of the weighted distances is minimal.

See for a Peyré et al. (2019) more detailed explanation.

Since our datasets are smaller than typical image classification datasets, we can afford to compute the optimal transport between the complete training set, and a 3 times larger sample of generated data.

## Network architecture

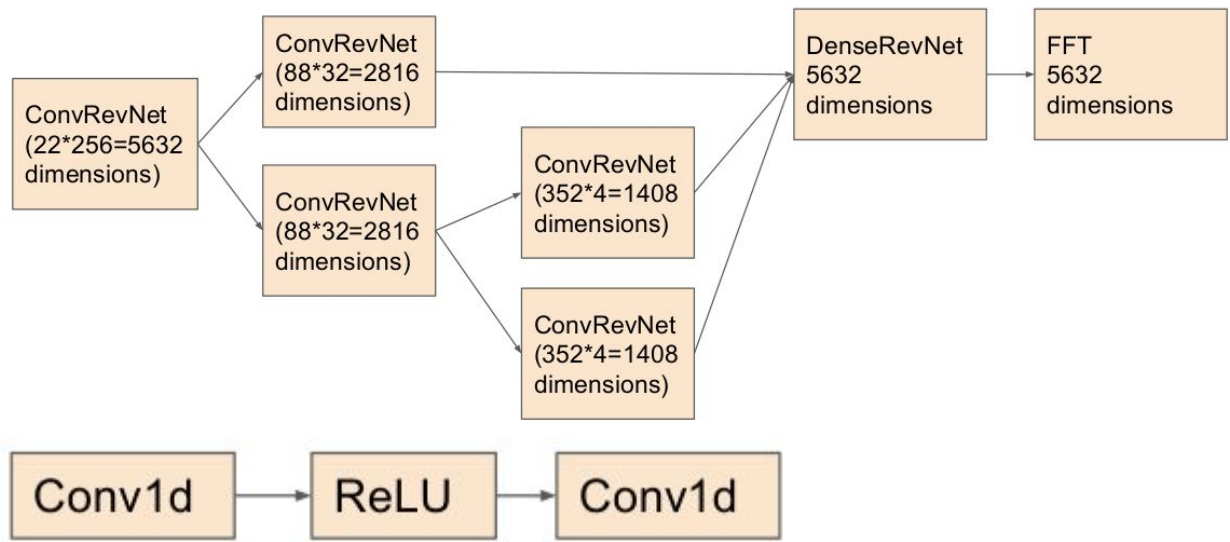

**Figure 1: Network Architecture.** ConvRevNet here always consists of a chain of several subchains of: 1 Subsampling Layer, 2 Additive Coupling Blocks with F and G functions as shown below (Conv1d, ReLU, Conv1D), 1 Subsampling Layer,... We include an FFT layer at the very end for easier aggregation of global information.

Our concrete architecture consists of subsampling layers and additive coupling layers and is schematically shown in Figure 1.

## High Gamma dataset

The High-Gamma dataset contains 4 second trials of either executed right hand, left hand, feet movements or of resting state. It is described in more detail (Schirrmeister et. al, 2017a). In this manuscript, we only use the right hand and resting state trials as they should yield very different types of signals. Also, since we are still developing the model, we restricted ourselves to evaluations on a 80/20 split on the training set and did not include the final evaluation dataset in any form.

# Results

|  | Test accuracy [%] | Train accuracy [%] |
|---|---|---|
| Invertible Network | 82.9 | 89.1 |
| Deep ConvNet | 92.5 | 100 |
| Shallow ConvNet | 93.0 | 100 |

**Table 1: Decoding accuracies from optimal transport-based optimization.** Results on subjects 5-9 from the High-Gamma dataset.

## Optimal transport

Optimal transport-based optimization yielded decoding accuracies substantially above chance (50%), however also roughly 10% worse than conventional convolutional networks (see Table 1). Also, the deep ConvNet trained purely on generated signals reaches ~64% accuracy on real signals, showing that the generated signals do retain some of the discriminative information from the real signals.
Class-conditional means in the output-space represented meaningful representative class-prototypes when visualized in the input space as seen in Figure 2. Individual dimensions are harder to interpret (Figure 3).

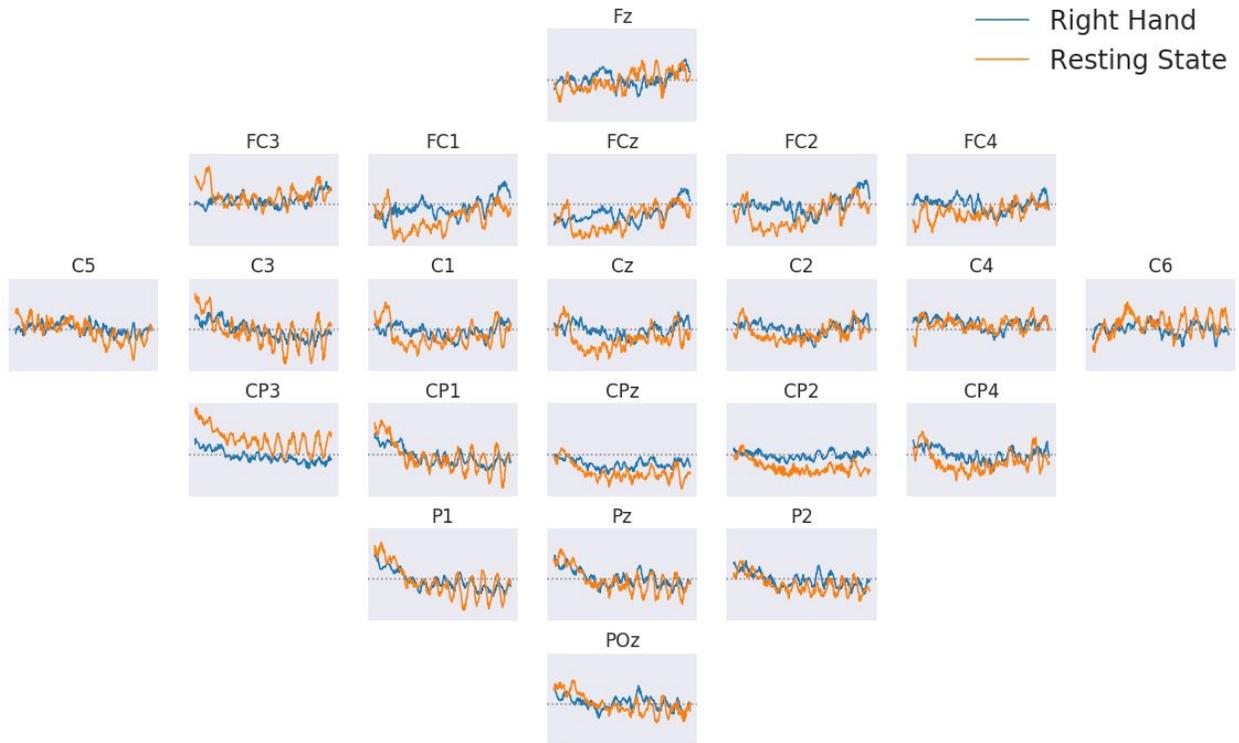

**Figure 2: Class-conditional output-space means visualized in the input space.** Signals show the result of inverting the optimized class means in the output space to the input space via the invertible network for an exemplary subject. Typical expected class differences between right hand and resting state visible such as suppressed alpha rhythms on the left side for the right hand movement, however also unexpected differences such as baseline differences between the classes visible.

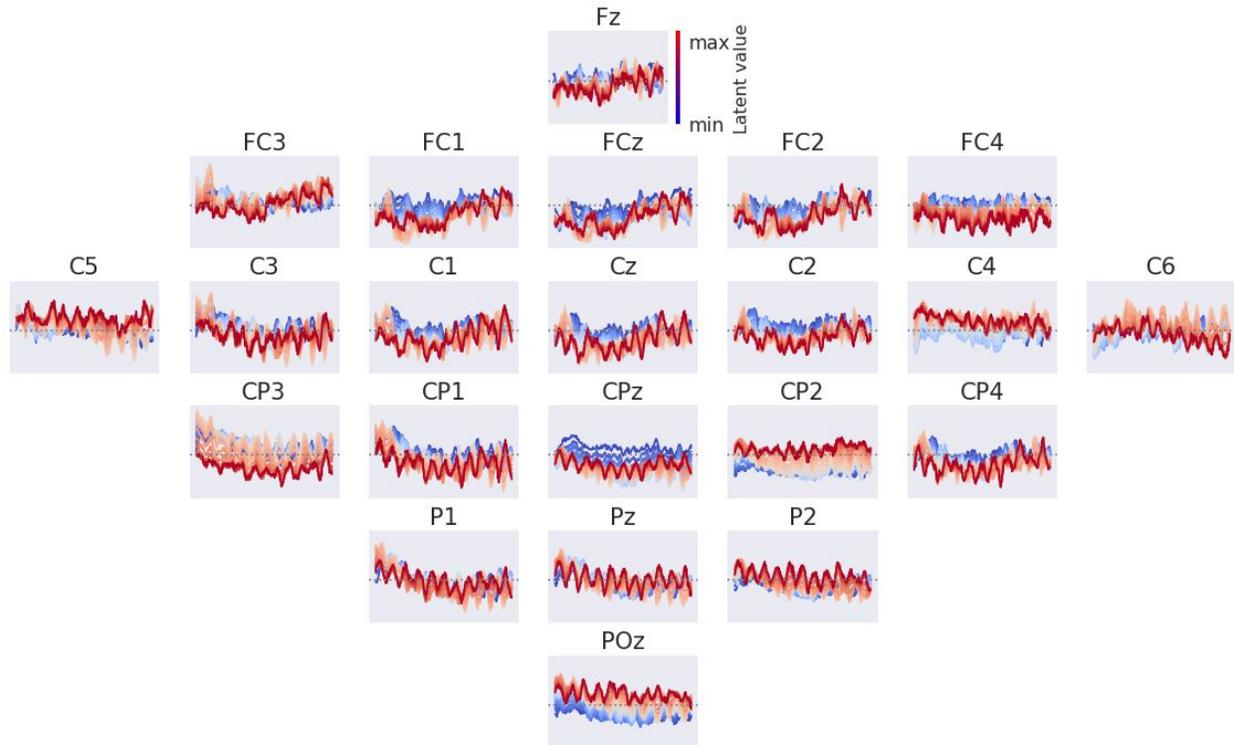

**Figure 3: Individual dimensions of the output space of an invertible network.** Starting at the mean for one class, changing one dimension in the out space from a low (blue) to a high (red) value and visualizing the correspondingly inverted signals in the input space.

## Maximum likelihood

Also with maximum likelihood, we were able to generate realistic EEG-signals (see Figure 4), both as judged by their visual appearance, as well as when comparing the spectra of generated and real EEG signals (Figure 5). In this preliminary work and without a lot of tuning, we were not able to reach more than 82% accuracy on a representative subject, where conventional convolutional networks can reach above 90% accuracy, and therefore did not evaluate it on all subjects as a classification model yet.

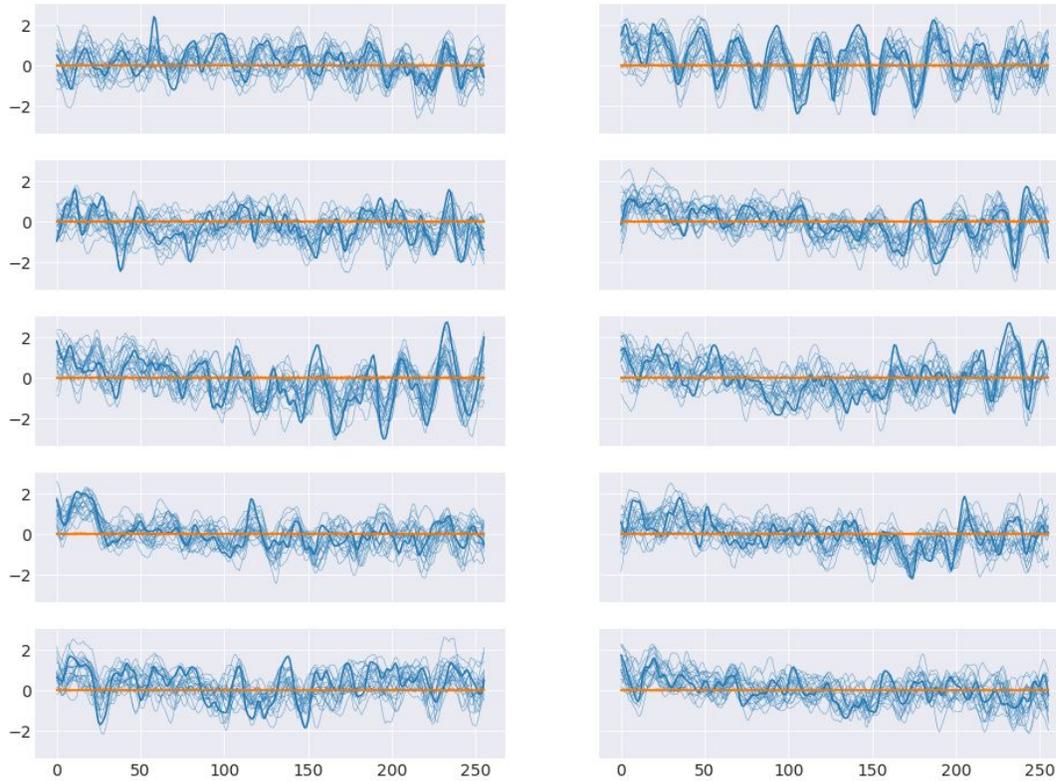

**Figure 4: Examples of matched generated and real data from the maximum likelihood trained model.** Thick lines represent real data, thin lines generated data samples, matching of real and generated data points by optimal transport. Blue lines show a real EEG channel, orange lines show a virtual channel that is always zero, existing for implementation reasons. Generated data often closely follows real data, for real channel as well as for virtual channel.

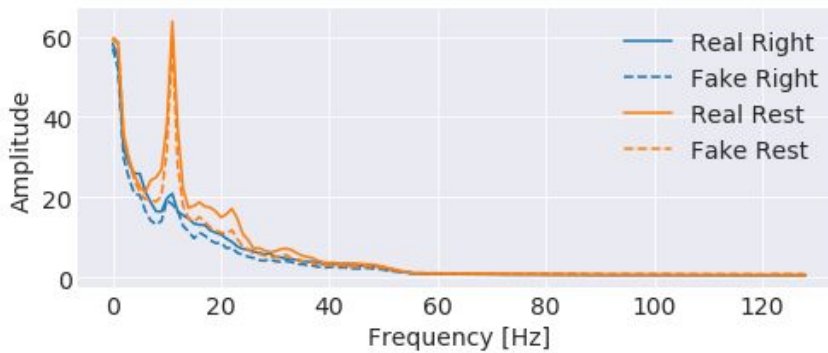

**Figure 5: Spectral of real and generated data from the maximum likelihood trained model.** Spectra are qualitatvely very similar between generated and real signals.

# Unevaluated Ideas

For the regularization of the invertible networks, we developed two ideas which we have not properly evaluated yet. The first idea is to use invertible networks to perform minibatch-based mixture model estimation with one mixture per datapoint, the second idea is to compare likelihoods of the trained invertible network on a training and a validation dataset to likelihoods of a fixed simple prior distribution and use these to dynamically regularize the network.

## Invertible networks for mixture model optimized using minibatches

Invertible networks can be used to scale up mixture model estimation with one mixture per datapoint to large datasets using minibatches. Normally to compute the likelihood of a datapoint under a mixture model, one has to have access to each mixture component since the likelihood is the average likelihood across all mixture components (assuming a uniform distribution over the mixture components). This can make such likelihoods computationally very expensive to compute in case one wants to use each training datapoint as the center of a mixture component, e.g., of a gaussian distribution. However, one can also train an invertible network using maximum likelihood optimization to learn the mixture distribution. Crucially, this now allows to optimize parameters of the mixture component distributions on a validation set using minibatches as follows. Assuming the invertible network $f$ currently computes the exact likelihoods of the mixture distribution and given a batch of training and validation datapoints, one can first subtract the likelihoods of the corresponding training mixture components from the invertible network likelihoods on the validation datapoints. This way, one can for example train individual standard deviations for each training point mixture component.

## Regularization by Comparison to Prior Distribution

We also developed an idea to regularize the invertible network dynamically during training as follows. Given a splitting into training and validation data, track the average log likelihood of the invertible network and of a simple prior such as a gaussian in the input space both for the training and the validation data. Whenever the increase in log likelihood from the gaussian prior to the current invertible network for the validation data becomes smaller then the increase in log likelihood of the invertible model between the validation and the training data, optimize the invertible network towards the likelihoods from the simple prior.

# Conclusion

We have developed an deep invertible network architecture that can simultaneously classify and generate EEG signals. We have suggested two ideas to further regularize them so that their decoding accuracies may become more competitive in comparison with conventional convolutional networks for EEG decoding.